\documentclass[10pt,twocolumn,letterpaper]{article}

\usepackage{cvpr}
\usepackage{times}
\usepackage{epsfig}
\usepackage{graphicx}
\usepackage{amsmath}
\usepackage{amssymb}
\usepackage[table]{xcolor}
\usepackage{bm}
\usepackage{subfigure}
\usepackage{color}
\usepackage{multirow}

\usepackage[pagebackref=true,breaklinks=true,letterpaper=true,colorlinks,bookmarks=false]{hyperref}
\DeclareMathOperator*{\argmin}{argmin}
\DeclareMathOperator*{\argmax}{argmax}

\cvprfinalcopy 

\def\cvprPaperID{152} 

\begin{document}

\title{Extreme Channel Prior Embedded Network for Dynamic Scene Deblurring}

\author{Jianrui Cai$^{1}$, Wangmeng Zuo$^{2}$, Lei Zhang$^{1}$\\
$^{1}$Department of Computing, The Hong Kong Polytechnic University, Hong Kong, China\\
$^{2}$School of Computer Science and Technology, Harbin Institute of Technology, Harbin, China\\
{\tt\small \{csjcai, cslzhang\}@comp.polyu.edu.hk, wmzuo@hit.edu.cn}
}

\maketitle
\begin{abstract}
Recent years have witnessed the significant progress on convolutional neural networks (CNNs) in dynamic scene deblurring.
While CNN models are generally learned by the reconstruction loss defined on training data, incorporating suitable image priors as well as regularization terms into the network architecture could boost the deblurring performance. 
In this work, we propose an Extreme Channel Prior embedded Network (ECPeNet) to plug the extreme channel priors (\ie, priors on dark and bright channels) into a network architecture for effective dynamic scene deblurring.
A novel trainable extreme channel prior embedded layer (ECPeL) is developed to aggregate both extreme channel and blurry image representations, and sparse regularization is introduced to regularize the ECPeNet model learning. 
Furthermore, we present an effective multi-scale network architecture that works in both coarse-to-fine and fine-to-coarse manners for better exploiting information flow across scales.
Experimental results on GoPro and K{\"o}hler datasets show that our proposed ECPeNet performs favorably against state-of-the-art deep image deblurring methods in terms of both quantitative metrics and visual quality.
\end{abstract}
\begin{figure*}
\footnotesize
\centering
\subfigure{
\begin{minipage}{0.333\textwidth}
\centering
\includegraphics[width=1\textwidth]{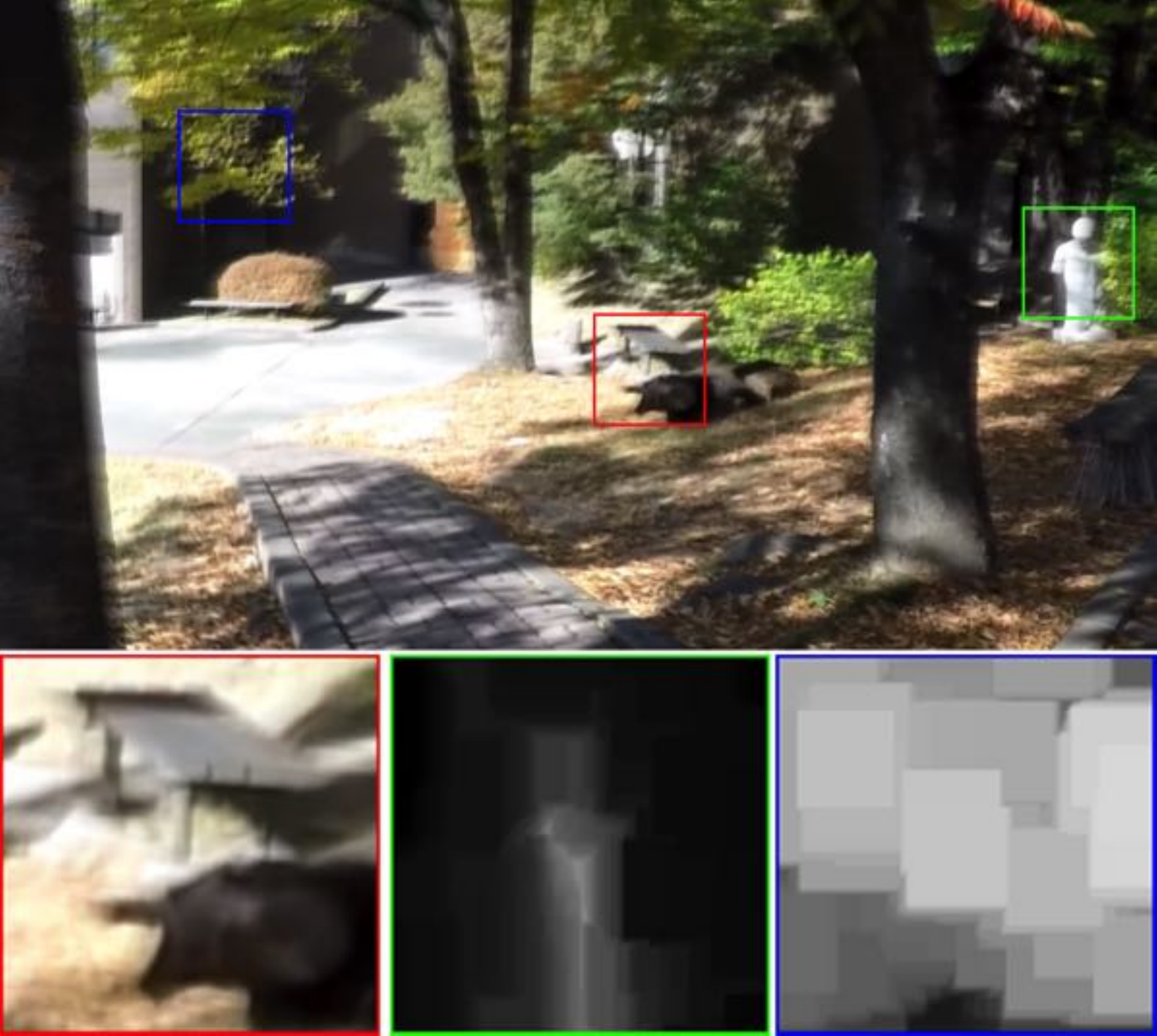}
{(a) Blurry image}
\end{minipage}
\begin{minipage}{0.333\textwidth}
\centering
\includegraphics[width=1\textwidth]{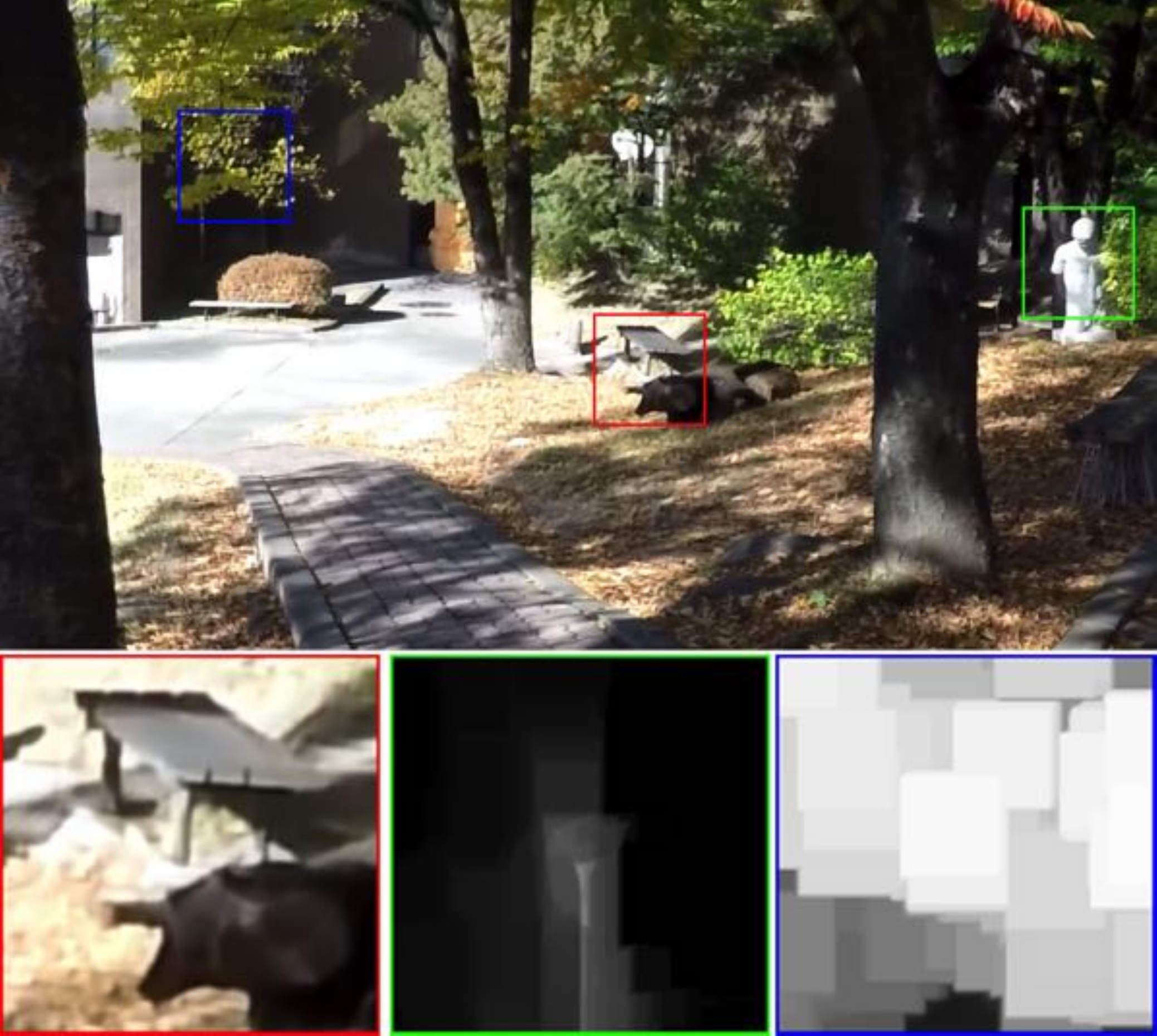}
{(b) Our result}
\end{minipage}
\begin{minipage}{0.333\textwidth}
\centering
\includegraphics[width=1\textwidth]{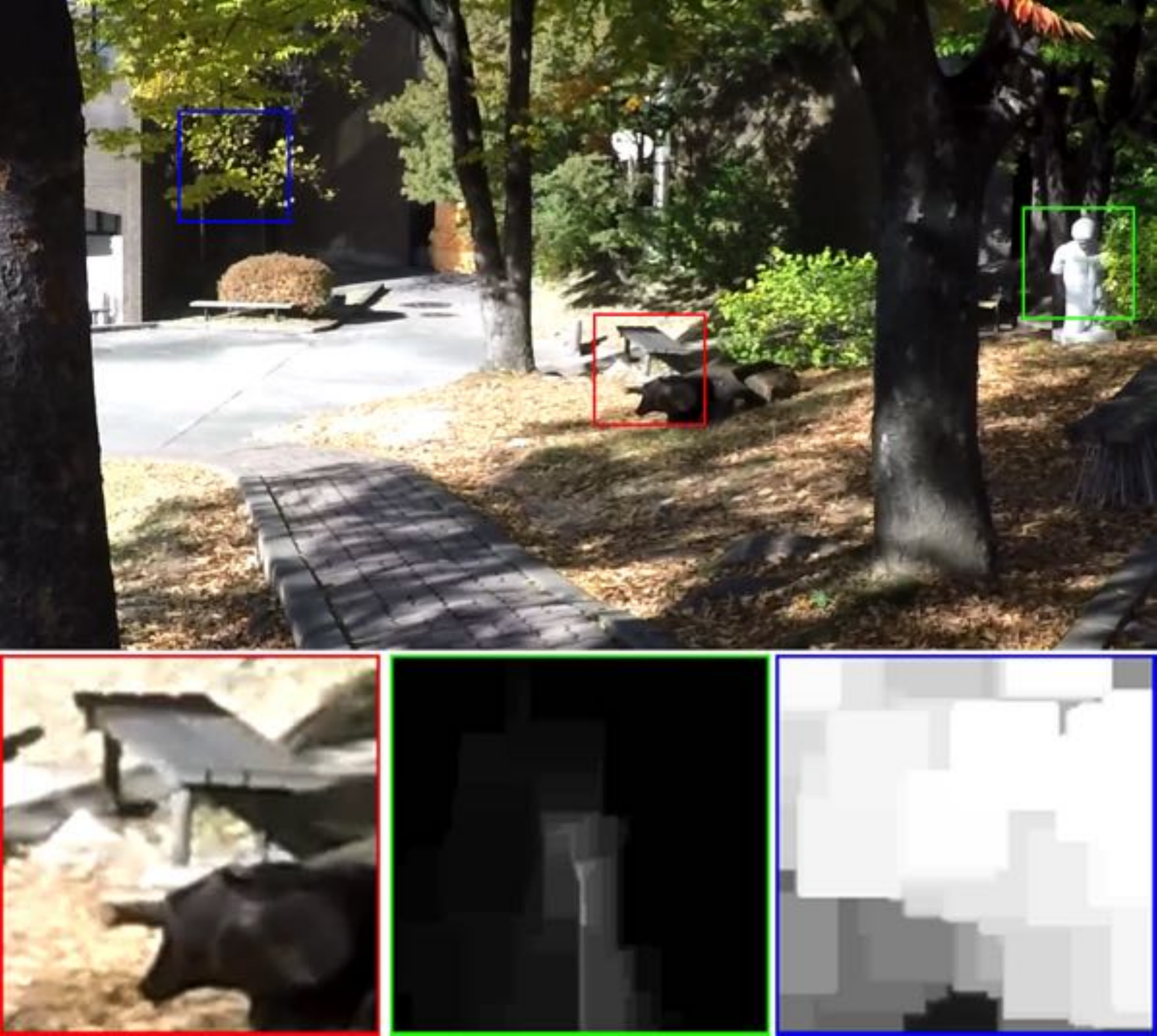}
{(c) Ground truth}
\end{minipage}
}
\caption{Deblurring result on GoPro image \cite{nah2017deep}. {\color{red}\textbf{Red box:}} zoom-in view of the original local patch. {\color{green}\textbf{Green box:}} zoom-in view of the dark channel of its corresponding local patch. {\color{blue}\textbf{Blue box:}} zoom-in view of the bright channel of its corresponding local patch.}
\label{fig:Intro}
\end{figure*}
\section{Introduction}
Reproducing a high quality image faithful to the scene is an essential goal of digital photography.
The real images, however, are often blurred during image acquisition due to the effect of many factors such as camera shake, object motion, and out-of-focus \cite{nah2017deep}.
The resulting blurry images will not only degrade the perceptual quality of photos but also degenerate the performance of many image analytic and understanding models \cite{kupyn2017deblurgan}.
Blind image deblurring, which has been studied extensively in low level vision for decades of years \cite{lucy1974iterative}, plays an essential role in improving the visual quality of real-world blurry images.
In general, the purpose of blind image deblurring is to recover the latent sharp image $\textbf{y}$ from its blurry observation $\textbf{x}$ = $\textbf{k}$ $\otimes$ $\textbf{y}$ + $\textbf{n}$, where $\textbf{k}$ is an unknown blur kernel (\ie, uniform or non-uniform), \textbf{n} is an additive white Gaussian noise and $\otimes$ denotes the convolution operator.
This inverse problem, however, is severely ill-posed and requires extra information on latent image $\textbf{y}$ to constrain the solution space.
Thus, there are two categories of approaches for utilizing prior knowledge, \ie, optimization-based and deep learning based deblurring methods.
Optimization-based approaches explicitly model prior knowledge to regularize the solution space of blur kernel \cite{chan1998total, krishnan2011blind, pan2014deblurring, sun2013edge, zuo2015discriminative} and latent image \cite{fergus2006removing, bahat2017non, pan2016blind, yan2017image, li2018learning}.
In contrast, deep learning based methods \cite{nah2017deep,kupyn2017deblurgan, tao2018scale, zhang2018dynamic} implicitly utilize prior knowledge by learning a direct mapping (\eg, convolutional neural network, CNN) from degraded image to latent clean image.

For blind image deblurring problem, optimization-based and deep learning methods respectively have their merits and limitations.
Optimization-based methods are flexible in incorporating versatile prior or regularization \cite{chan1998total, krishnan2011blind, zuo2015discriminative, pan2016blind, yan2017image} tailored for blind deblurring, but suffer from the time-consuming optimization procedure and over-simplified assumptions on blur kernel (\eg, spatially invariant and uniform).
Moreover, conventional image priors (\eg, total variation~\cite{chan1998total}) are limited in blind deblurring and prone to ordinary solution of delta kernel.
Specific priors, \eg, $\ell_{0}$-norm \cite{xu2013unnatural} and normalized sparsity \cite{krishnan2011blind}, are then suggested for blur kernel estimation.
On the other hand, deep learning methods \cite{nah2017deep, kupyn2017deblurgan, tao2018scale, zhang2018dynamic} benefiting from end-to-end training and joint optimization can enjoy a fast speed and flexibility in handling spatially variant blur in the dynamic scene.
However, deep models learn the direct mapping for blind deblurring, and may be limited in capturing specific priors for blind deblurring.
As for dynamic scene deblurring, existing dataset \cite{nah2017deep} is of relatively small scale, which may be a factor hindering the performance of learned model.
Taking the merits and drawbacks of optimization-based and deep learning based methods into account, one interesting question is to ask whether we can exploit prior model to constrain both the network architecture and learning losses for improved dynamic scene deblurring performance.
In this paper, we make the first attempt to address this challenging problem.
In particular, based on the effectiveness of image prior in blind deblurring, we propose an Extreme Channel Prior embedded Network (ECPeNet) to help the restoration of latent clean image.
The critical component of ECPeNet is a novel trainable extreme channel prior embedded layer (ECPeL), which can aggregate extreme channel and blurry image representations to leverage their respective advantages.
By enforcing sparsity on both dark and bright channel of feature maps, we can regularize the solution space of CNN during training, thereby incorporating extreme channel priors into ECPeNet.
Furthermore, existing deep dynamic scene deblurring models \cite{nah2017deep, tao2018scale} usually adopt multi-scale network architecture but only consider the coarse-to-fine information flow.
That is, blind deblurring is first performed at the small scale, and then deblurring results (or latent representations) are combined with feature representations of a larger scale for further refinement.
However, we show that feature representations of a larger scale actually also benefit the dynamic scene deblurring at a smaller scale.
To this end, we present a more effective multi-scale network architecture that works in both coarse-to-fine and fine-to-coarse manners for better exploiting information flow across scales.
Experimental results on GoPro and K{\"o}hler datasets show that our proposed ECPeNet performs favorably against state-of-the-art deep models.
With the above claimed advantages, the ECPeNet can produce visually pleasing results, as shown in Figure \ref{fig:Intro}.
The contribution of this paper is three-fold:
(i) We propose a novel trainable structural layer, which can aggregate both data information and prior knowledge together to leverage their respective merits but avoid limitations.
To the best of our knowledge, this is the first attempt to plug prior knowledge (\ie, statistical properties) into a deblurring network in an end-to-end manner.
(ii) We introduce a new multi-scale network architecture to fully exploit different resolution images for maximizing the information flow.
(iii) Extensive experimental results on GoPro \cite{nah2017deep} and K{\"o}hler \cite{kohler2012recording} datasets demonstrate that our ECPeNet outperforms state-of-the-art dynamic scene deblurring methods.
\section{Related Work}
In this section, we briefly review the recent optimization-based and deep learning based image deblurring methods.
\subsection{Optimization-based Deblurring Methods}
The optimization-based methods aim to develop effective image priors to favor clean images over the blurry one.
Representative priors include sparse gradients \cite{fergus2006removing, shan2008high, levin2009understanding, xu2010two}, hyper-Laplacian prior \cite{Krishnan09}, normalized sparsity prior \cite{krishnan2011blind}, L$_{0}$-norm prior \cite{xu2013unnatural}, patch recurrence prior \cite{michaeli2014blind} and discriminative learned prior \cite{zuo2015discriminative, li2018learning}.
By taking advantages of the aforementioned priors, existing optimization-based methods could deliver competitive results on generic natural images. 
These approaches, however, cannot be generalized well to handle domain specific images.
Thus, specific priors are needed to be introduced for specific images (\eg, a light streak prior \cite{hu2014deblurring} for low light images, and a combination of intensity and gradient prior \cite{pan2017l_0} for text images).
Recently, Pan \etal \cite{pan2016blind} developed a dark channel prior (DCP) \cite{he2011single} based model to enforce sparsity on the dark channel of latent image and achieved promising result on both generic and specific images.
With the success of \cite{pan2016blind}, Yan \etal \cite{yan2017image} further introduced a bright channel prior (BCP) to solve the corner case image, which contains a large amounts of bright pixels.
By plugging the extreme channel prior (a combination of BCP and DCP) into the deblurring model, Yan \etal achieved state-of-the-art results on various scenarios.
Although those algorithms demonstrate their effectiveness in image deblurring, the simplified assumptions on the blur model and time-consuming parameter-tuning strategy are two lethal problems to hinder their performance in real-world cases.
In this work, we utilize a realistic GoPro dataset \cite{nah2017deep} to end-to-end train a new multi-scale network for latent sharp image restoration.
\subsection{Deep Learning based Deblurring Methods}
Deep learning based methods focus on exploiting external training data to learn a mapping function accords with the degradation process.
The powerful end-to-end training paradigm and non-linear modeling capability make CNNs a promising approach to image deblurring.
Early CNN-based deblurring methods aim to mimic conventional deblurring frameworks for the estimation of both latent image and blur kernel.
Works in \cite{sun2015learning} and \cite{gong2017motion} first used networks to predict the non-uniform blur kernel and then utilized a non-blind deblurring method \cite{zoran2011learning} to restore images.
Schuler \etal \cite{schuler2016learning} introduced a two-stages network to simulate iterative optimization. 
Chakrabarti \etal \cite{chakrabarti2016neural} utilized a network to predict frequency coefficients of blur kernel.
However, these methods may fail when the estimated kernel is inaccurate \cite{ren2018partial}. 
Therefore, more recent approaches preferred to train kernel estimation-free networks to restore latent images directly.
Specifically, Nah \etal \cite{nah2017deep} proposed a multi-scale based CNN to progressively recover the latent image.
Tao \etal \cite{tao2018scale} introduced a scale-recurrent network which equipped with a ConvLSTM layer \cite{xingjian2015convolutional} to further ensure information flow between different resolution images.
Kupyn \etal \cite{kupyn2017deblurgan} adopted Wasserstein GAN \cite{goodfellow2014generative, arjovsky2017wasserstein} as an objective function to restore the texture details of latent image.
Zhang \etal \cite{zhang2018dynamic} employed spatially variant recurrent neural networks (RNNs) to reduce the computational cost.
While existing deblurring networks have reported impressive results, the limited number of training data and the disappreciation of prior knowledge may become two main factors hampering the performance improvement.
To mitigate aforementioned issues, we in this paper introduce the extreme channel prior for CNN and encourage it to constrain the solution space under these priors. 
\section{Extreme Channel Prior Embedded Network}
Given a single blurry image $\bm{x_i}$, existing CNN-based methods aim at learning a mapping $F_{\Theta}$ to generate an estimation of latent sharp image $\bm{\hat{y_i}}$, which is required to approximate the ground-truth $\bm{y_i}$.
This procedure can be formulated as:
\begin{equation}
\begin{aligned}
\hat{\Theta} = {\mathop{\argmin}_{\Theta}}{\sum\nolimits_{i}} \ell(\bm{\hat{y_i}}, \bm{y_i})
~~~~~~s.t.~~~\bm{\hat{y_i}} = F_{\Theta}(\bm{x_i})
\end{aligned}
\label{eqn:tradi}
\end{equation}
where (\bm{$x_i$}, \bm{$y_i$}) refer to the $\bm{i}$-th image pairs in the training dataset and $\Theta$ is the parameter of mapping function.

However, such formulation is limited in capturing image priors specified to blind deblurring, which is generally much different from those for non-blind restoration.
Moreover, the existing training image pairs are insufficient to learn an effective mapping function $F_{\Theta}$.
Therefore, we propose ECPeNet that aggregates both blurry image and extreme channel representations to enhance deblurring performance,
and Eqn. (\ref{eqn:tradi}) can be rewritten as:
\begin{equation}
\begin{aligned}
\hat{\Theta} = {\mathop{\argmin}_{\Theta}}{\sum\nolimits_{i}} \ell(\bm{\hat{y_i}}, \bm{y_i})
~~~~~s.t.~~~\bm{\hat{y_i}} = F_{\Theta}(\bm{x_i} | \Lambda, \Omega)
\end{aligned}
\label{eqn:our}
\end{equation}
where $\Lambda$ and $\Omega$ are the extreme channel representations under the constraint of both dark and bright channels priors.
By this way, we can embed the image priors specified to blind deblurring into the mapping function $F_{\Theta}$, and expect it to have ability to generate the higher quality latent sharp image.
In the following subsections, we present in detail the proposed ECPeNet.
%

\begin{figure*}
\begin{center}
\includegraphics[width=1\linewidth]{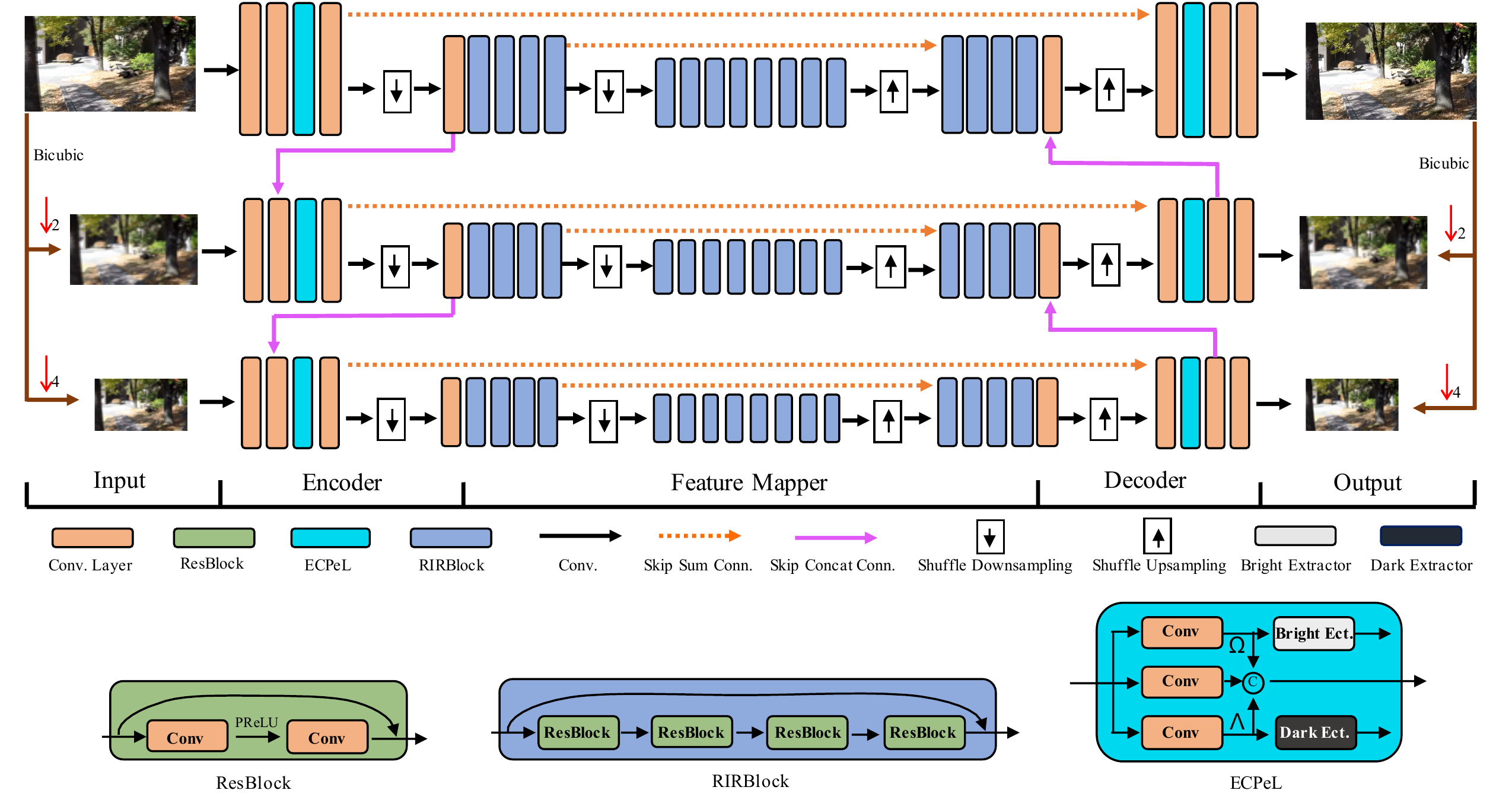}
\end{center}
\caption{Illustration of our proposed ECPeNet architecture.}
\label{fig:Arch}
\end{figure*}

%
\subsection{Architecture}
The overall architecture of our proposed ECPeNet is illustrated in Figure \ref{fig:Arch}.
It contains three sub-networks respectively for three scales, and each of them consists of three major components: (i) input and output; (ii) encoder and decoder; (iii) feature mapper.
Note that instead of utilizing Rectifier Linear Units (ReLU) \cite{krizhevsky2012imagenet}, we take parametric ReLU (PReLU) \cite{he2015delving} as activation function
since it can improve the modeling capability with negligible extra computational cost.
Unless denoted otherwise, all the convolution filters are set to $3\times3$, not $5\times5$ that utilized by most of the other dynamic scene deblurring networks (\eg, \cite{nah2017deep} and \cite{tao2018scale}).
Although a filter of size $5\times5$ has more parameters than two filters of size $3\times3$, the one utilizing $3\times3$ filters is more efficient since additional nonlinearity can be inserted between them \cite{simonyan2015very}.
Besides, the stride size for all convolution layers is set to $1$ and the number of feature maps in each layer is set to $64$, except for the last layer and the proposed ECPeL which are set to $3$ and $\{3, 64, 3\}$, respectively.
The details of each component are described as follows.
\vspace{0.8mm}
\noindent \textbf{Input and Output.}
An effective multi-scale network architecture is utilized in this work to restore the latent sharp image from a coarser scale to finer scale.
Consider an image pair $(\bm{x}, \bm{y}) \in \mathbb{R}^{H \times W \times C}$, where $H\times W$ is the pixel resolution and $C$ is the number of channels, which is set to $3$.
We first use bicubic interpolation to progressively downsample the image pair with the ratio of $\frac{1}{2}$, and generate $3$ scales image pairs with the resolution of $\{H\times W, \frac{H}{2}\times \frac{W}{2}, \frac{H}{4}\times \frac{W}{4}\}$.
We then take the blurred image at each scales as input to produce its corresponding sharp images.
The sharp one at original resolution is considered as the final output.
\vspace{0.8mm}
\noindent \textbf{Encoder and Decoder.}
Each scale of encoder consists of $4$ convolution layers, the proposed ECPeL and a shuffle operation with factor $\frac{1}{2}$.
As for decoders, they basically mirror the architecture of the encoders, except for the factor of shuffle operation \cite{shi2016real} is set to $2$.
The encoder and decoder networks are mainly designed for three purposes.
Firstly, they progressively transform different scale images to extract shallow features (in encoders) and transform them back to the resolution of inputs (in decoders).
Secondly, they downsample and upsample the shallow features to ensure information flow between different scale images and expand the receptive field.
Note that instead of directly concatenating the upsampled coarser scale latent image with the finer scale blurry image \cite{nah2017deep, chen2017photographic, tao2018scale}, we argue that
(i) the information of finer scale blurry image is beneficial for estimating the coarser scale latent image
and (ii) the concatenation in shallow feature domain can yield a better result than in image (RGB) domain.
Thus, the encoder first shuffles features with factor $\frac{1}{2}$ to ensure the same resolution between different scales features, \eg, a features with size $m\times n\times c$ is shuffled to $\frac{m}{2}\times \frac{n}{2}\times 4c$.
Then it concatenates the downsampled features with coarser scale features for restoring the coarser scale sharp image.
On the contrary, the decoder shuffles features with size $\frac{m}{2}\times \frac{n}{2}\times 4c$ back to $m\times n\times c$ and concatenates them with finer scale features for predicting the finer scale sharp image.
Benefiting from this multi-scale architecture, the network can fully exploit different scale images to maximize the information flow between them, resulting in better performance.
In Section \ref{sec:as}, we conduct an ablation study to verify its effectiveness.
Thirdly, they integrate the extreme channel prior into the network via ECPeL.
We provide more details of the proposed ECPeL in Section \ref{sec:ecpel}.
\vspace{0.8mm}
\noindent \textbf{Feature Mapper.}
The feature mapper module, which aims to refine the shallow features progressively, is an essential part of the latent image restoration.
One critical factor for reducing blur artifacts is the size of receptive field.
To enlarge the receptive field, we (i) stack a set of convolutional layers to achieve a larger depth network and (ii) utilize the shuffle operations for downsampling and upsampling the features.
Considering that a deeper neural network is more challenging to converge, we adopt the residual blocks to speed-up the training procedure \cite{he2016deep, kim2016accurate}.
Besides, the feature mapper module utilizes the long skip connection and short skip connection to make full use of hierarchical features in all convolutional layers.
A similar skip connection strategy has been utilized in a very recent work \cite{zhang2018image}.
As illustrated in Figure \ref{fig:Arch}, the feature mapper contains $16$ residual in residual blocks (RIRBlock), and each of them has $4$ residual blocks (ResBlock).
The ResBlock consists of $2$ convolution layers and a PReLU activation function.
Since we utilize the filter of size $3\times3$, the total parameters of our ECPeNet are almost in the same magnitude as the previous methods.
Note that all the weights across different scales feature mapper sub-modules are shared.
\subsection{Extreme Channel Prior Embedded Layer}\label{sec:ecpel}
The proposed ECPeL is designed for aggregating both blurry image representation and extreme channel representation to regularize the solution space of CNN.
Specifically, it first learns $3$ mapping functions $\mathcal{M}_{\theta}$, $\mathcal{M_{[\alpha|D]}}$ and $\mathcal{M}_{[\beta|B]}$ to transform the feature map $f^{l-1}$ from previous layer into $3$ new feature maps, including a deeper layer transformed feature $f^{l}$, a dark channel prior constrained feature $\Lambda$, and a bright channel prior constrained feature $\Omega$.
It then adopts a concatenation operation to concatenate those $3$ feature maps for the integration of blurry image representation and extreme channel representation.
Formally, the proposed ECPeL can be expressed as:
\begin{equation}
\begin{aligned}
\lbrack\Lambda, f^{l}, \Omega\rbrack = \textit{ECPeL}(f^{l-1}) \\
f^{l} = \mathcal{M}_{\theta}(f^{l-1})\\
\Lambda = \mathcal{M}_{[\alpha|D]}(f^{l-1})\\
\Omega = \mathcal{M}_{[\beta|B]}(f^{l-1})
\end{aligned}
\label{eqn:ecpel}
\end{equation}
where $\lbrack\Lambda, f^{l}, \Omega\rbrack$ denotes the concatenation of the feature maps, and the subscripts $[\alpha|D]$ and $[\beta|B]$ denote that parameters $\alpha$ and $\beta$ are optimized under the dark and bright channel prior constraint.
To add extreme channel prior constraint into a network, the ECPeL utilizes (i) extractors to extract both dark and bright channel of features, and (ii) the $\ell_1$-regularization term to enforce sparsity in training.
The extractor $D(\bm{\cdot})$ is designed to extract the dark channel of $\Lambda$ via computing its minimum values in a local patch.
Formally, its $\texttt{forwards}$ function can be written as follows:
\begin{equation}
\begin{aligned}
D(\Lambda)&_{[h, w]} = \Lambda_{[\mathcal{I_D}_{[h, w]}]}\\
\mathcal{I_D}_{[h, w]} = &~{\mathop{\argmin}}_{i^{\star} \in \Psi_{[h, w, c]}}\Lambda_{[i^{\star}]}
\end{aligned}
\label{eqn:dforward}
\end{equation}
The extractor $B(\bm{\cdot})$ aims to extract the bright channel of  $\Omega$ by calculating its maximum values in a local patch.
And its $\texttt{forwards}$ function can be formulated as:
\begin{equation}
\begin{aligned}
B(\Omega)&_{[h, w]} = \Omega_{[\mathcal{I_B}_{[h, w]}]}\\
\mathcal{I_B}_{[h, w]} = &~{\mathop{\argmax}}_{i^{\star} \in \Psi_{[h, w, c]}}\Omega_{[i^{\star}]}
\end{aligned}
\label{eqn:bforward}
\end{equation}
where $\Psi_{[h, w, c]}$ is the index set of inputs in a sub-window centered at a pixel location $[h, w, c]$, $\mathcal{I_D}_{[h, w]}$ and $\mathcal{I_B}_{[h, w]}$ are the masks that records an index of the minimum and maximum value in a local patch, respectively.
The patch sizes for each scale are set to $\{31\times31, 19\times19, 11\times11\}$.
A single element $\Lambda_{[h, w, c]}$ and $\Omega_{[h, w, c]}$ of the input may be assigned to several different outputs $D(\Lambda)_{[h, w]}$ and $B(\Omega)_{[h, w]}$.
The $\texttt{backwards}$ function of extractors computes partial derivative of the loss function with respect to each input variable $\Lambda_{i}$ and $\Omega_{i}$ as follows:
\begin{equation}
\begin{aligned}
&\frac{\partial{L}}{\partial{\Lambda_{i}}} = \sum_h\sum_w\sum_c1\{i = \mathcal{I_D}_{[h, w]}\} \frac{\partial{L}}{\partial{D(\Lambda)_{[h, w]}}}\\
&\frac{\partial{L}}{\partial{\Omega_{i}}} = \sum_h\sum_w\sum_c1\{i = \mathcal{I_B}_{[h, w]}\} \frac{\partial{L}}{\partial{B(\Omega)_{[h, w]}}}
\end{aligned}
\label{eqn:backward}
\end{equation}
where $i$ refers to the pixel location $[h, w, c]$.
In words, the partial derivatives $\frac{\partial{L}}{\partial{D(\Lambda)_{[h, w]}}}$ and $\frac{\partial{L}}{\partial{B(\Omega)_{[h, w]}}}$ are accumulated if $i$ is the $\argmin$ and $\argmax$ selected for $D(\Lambda)_{[h, w]}$ and $B(\Omega)_{[h, w]}$, respectively.
In back-propagation, the partial derivatives $\frac{\partial{L}}{\partial{D(\Lambda)_{[h, w]}}}$ and $\frac{\partial{L}}{\partial{B(\Omega)_{[h, w]}}}$ are already calculated by the $\texttt{backwards}$ function of the loss layer.
With the proposed ECPeL, we can extract the dark and bright channel of shallow features (\ie, $D(\Lambda)$ and $B(\Omega)$), which can be further enforced to be sparse via the objective function.
By integrating the constrained features $\Lambda$ and $\Omega$ into the network, the proposed ECPeNet can achieve a better performance while using the same training dataset.
The ablation study in Section \ref{sec:as} is conducted for the evaluation.
\subsection{Loss Function}
We utilize the $\ell_{1}$-norm of the reconstruction error as loss function for each scale.
More specifically, we can rewrite Eqn. (\ref{eqn:our}) as:
\begin{equation}
\begin{aligned}
\mathcal{L} = \frac{1}{N} \sum_{i=1}^{N}\sum_{j=1}^{3}\|\bm{y{_i^j}} - F_{\Theta}(\bm{x{_i^j}} | \Lambda^{\bm{j}}, \Omega^{\bm{j}})\|_{1}
\end{aligned}
\label{eqn:loss}
\end{equation}
where $N$ is the total number of training pairs (\bm{$x$}, \bm{$y$}) and $\bm{j}$ is the number of scales, which is set to $3$ in this paper.
$(\bm{\cdot})^{\bm{j}}$ is a symbol referring to the image and feature in the $\bm{j}$-th scale.
According to the observation by \cite{pan2016blind, yan2017image}, a dark channel of images would be less dark after the blurring process, while a bright channel of images would be no longer bright.
The reason is that dark/bright pixel would be averaged with its neighboring high/low intensity pixels during a blurring process.
The sparsity regularization term is thus more beneficial for restoring a sharp image than a blurred one.
To this end, we introduce a $\ell_1$-regularization term to enforce sparsity on both dark and bright channels of shallow features.
The objective function can be given by:
\begin{equation}
\begin{aligned}
\mathcal{L} = \frac{1}{N} \sum_{i=1}^{N}\sum_{j=1}^{3}&\|\bm{y{_i^j}} - F_{\Theta}(\bm{x{_i^j}} | \Lambda^{\bm{j}}, \Omega^{\bm{j}})\|_{1} \\
& + \lambda \|D(\Lambda^{\bm{j}})\|_1 + \omega \|1 - B(\Omega^{\bm{j}})\|_1
\end{aligned}
\label{eqn:lossl1}
\end{equation}
where $\lambda$ and $\omega$ are the trade-off parameters.
$D(\bm{\cdot})$ and $B(\bm{\cdot})$ are the extractors to extract the dark channel and bright channel of features, respectively.
With the $\texttt{forwards}$ and $\texttt{backwards}$ functions, the dark and bright channel extractors can be jointly end-to-end optimized with the network.
\section{Experimental Results}
In this section, we provide experimental results to show the advantage of our proposed ECPeNet.
We implement our framework by using Caffe toolbox \cite{jia2014caffe}, and train the model on a PC equipped with an Intel Core i7-7820X CPU, 128G RAM and a single Nvidia Quadro GV100 GPU.
\vspace{0.5mm}
\noindent \textbf{Datasets.}
We train our proposed ECPeNet on the GoPro training dataset \cite{nah2017deep}, which contains $22$ sequences with $2,103$ blurred/clear image pairs.
Once the model is trained, we test it on the standard GoPro testing dataset \cite{nah2017deep} and K{\"o}hler \cite{kohler2012recording} dataset.
The GoPro testing dataset consists of $11$ sequences with $1,111$ image pairs, and the K{\"o}hler dataset has 4 latent images and 12 blur kernels.
Note that, to simulate the realistic blurring process, the GoPro dataset generates blurred images through averaging adjacent short-exposure frames captured by a high-speed video camera, and the K{\"o}hler dataset replays the recorded $6D$ real camera motion trajectory to synthesize blurred images.
\vspace{0.5mm}
\noindent \textbf{Parameter Settings.}
We crop the GoPro training dataset (linear subset) into $256\times256\times3$ patches and make use of these patches to train the ECPeNet.
The mini-batch size in all the experiments is set to $10$, and the trade-off parameters $\lambda$ and $\omega$ are set to $0.1 (0.2)$ in this paper.
For the model training, we utilize Xavier \cite{glorot2010understanding} to initialize all trainable variables. 
The Adam solver \cite{kingma2014adam} is adopted to optimize the network parameters.
The default parameters of Adam solver are set as $\beta_1=0.9$, $\beta_2=0.999$ and $\epsilon=10^{-8}$.
We fix the learning rate as $10^{-4}$ and train the network with $600K$ iterations, which takes about $140$ hours.
Additionally, we randomly rotate and/or flip the image patches for data augmentation.
The $1\%$ additive Gaussian noise is also randomly added to the blurred images for robust learning.
\begin{table}
\small
\caption{Ablation study of extreme channel prior (ECP) constraint and image fully exploitation (IFE) strategy. The average PSNR (dB) on GoPro testing dataset with $150K$ iterations.}
\begin{center}
\begin{tabular}{|c||c|c|c|c|}
\hline
&\multicolumn{4}{|c|}{Different combinations of ECP and IFE}\\
\hline  \hline
ECP                  &$\times$   &$\surd$    &$\times$  &$\surd$\\
IFE                  &$\times$   &$\times$   &$\surd$   &$\surd$\\
\hline \hline
PSNR                 &$28.63$    &$28.86$    &$28.79$   &$28.95$\\
\hline
\end{tabular}
\end{center}
\label{tab:1}
\end{table}
\subsection{Ablation Study}\label{sec:as}
It is generally agreed that a larger scale training dataset which covers various image contents and blur models will bring benefit to train a robust deep network.
The type of scenes and number of images in the current GoPro dataset, however, are barely sufficient to train an efficient network. 
Rather than enlarging the training dataset, in this work, we propose to integrate the extreme channel prior into CNN and fully exploit different scales images for the performance improvement.
Here, we compare our network with several baseline models to verify the effectiveness of the extreme channel prior (ECP) constraint and image fully exploitation (IFE) strategy.
Table \ref{tab:1} shows the investigation on the effects of ECP constraint and IFE strategy.
It can be seen that the one without both ECP and IFE performs much worse than the ECPeNet in terms of PSNR ($28.63$ dB $v.s.$ $28.95$ dB).
By plugging the ECP constrain into the network, we can significantly improve the performance with a $0.23$ dB gain.
While compared to the networks that utilizes the multi-scale architectures \cite{nah2017deep, tao2018scale} or cascaded refinement \cite{chen2017photographic} strategy, the one adopting the proposed IFE strategy (both coarse-to-fine and fine-to-coarse manners) can has $0.16$ dB improvement in terms of PSNR index.
Note that we train all these $4$ networks with $150K$ iterations for the testing in this ablation study.
These comparisons firmly indicate the proposed ECP and IFE benefit for performance improvement.
\begin{table}
\small
\caption{Average PSNR (dB), SSIM, MSSIM indices and running time for different methods on the benchmark datasets (running time is measured for an image with the size of $1280\times720\times3$).}
\begin{center}
\begin{tabular}{|c|c|c|c|c|c|}
\hline 
\multirow{2}{*}{Method}  &\multicolumn{2}{c|}{GoPro} &\multicolumn{2}{c|}{K{\"o}hler} &\multirow{2}{*}{Time}\\
\cline{2-5}
                                            &PSNR   &SSIM   &PSNR   &MSSIM  &\multicolumn{1}{c|}{}\\
\hline \hline
Kim   \cite{hyun2014segmentation}           &23.64  &0.824  &24.68  &0.794   &$1$ hr      \\
\hline \hline
Sun   \cite{sun2015learning}                &24.64  &0.843  &25.22  &0.774   &$20$ min    \\
Nah   \cite{nah2017deep}                    &29.08  &0.914  &26.48  &0.808   &$2.87 s$    \\
Tao   \cite{tao2018scale}                   &30.26  &0.934  &26.75  &0.837   &$0.62 s$    \\
Kupyn \cite{kupyn2017deblurgan}             &28.70  &0.858  &26.10  &0.816   &$0.59 s$    \\
Zhang \cite{zhang2018dynamic}               &29.19  &0.931  &25.71  &0.800   &$0.76 s$    \\
\hline \hline
\rowcolor{gray!40} 
Proposed    								&31.10  &0.945  &26.79  &0.839   &$0.65 s$    \\
\hline
\end{tabular}
\end{center}
\label{tab:2}
\end{table}
%

\begin{figure*}
\begin{center}
\includegraphics[width=0.939\linewidth]{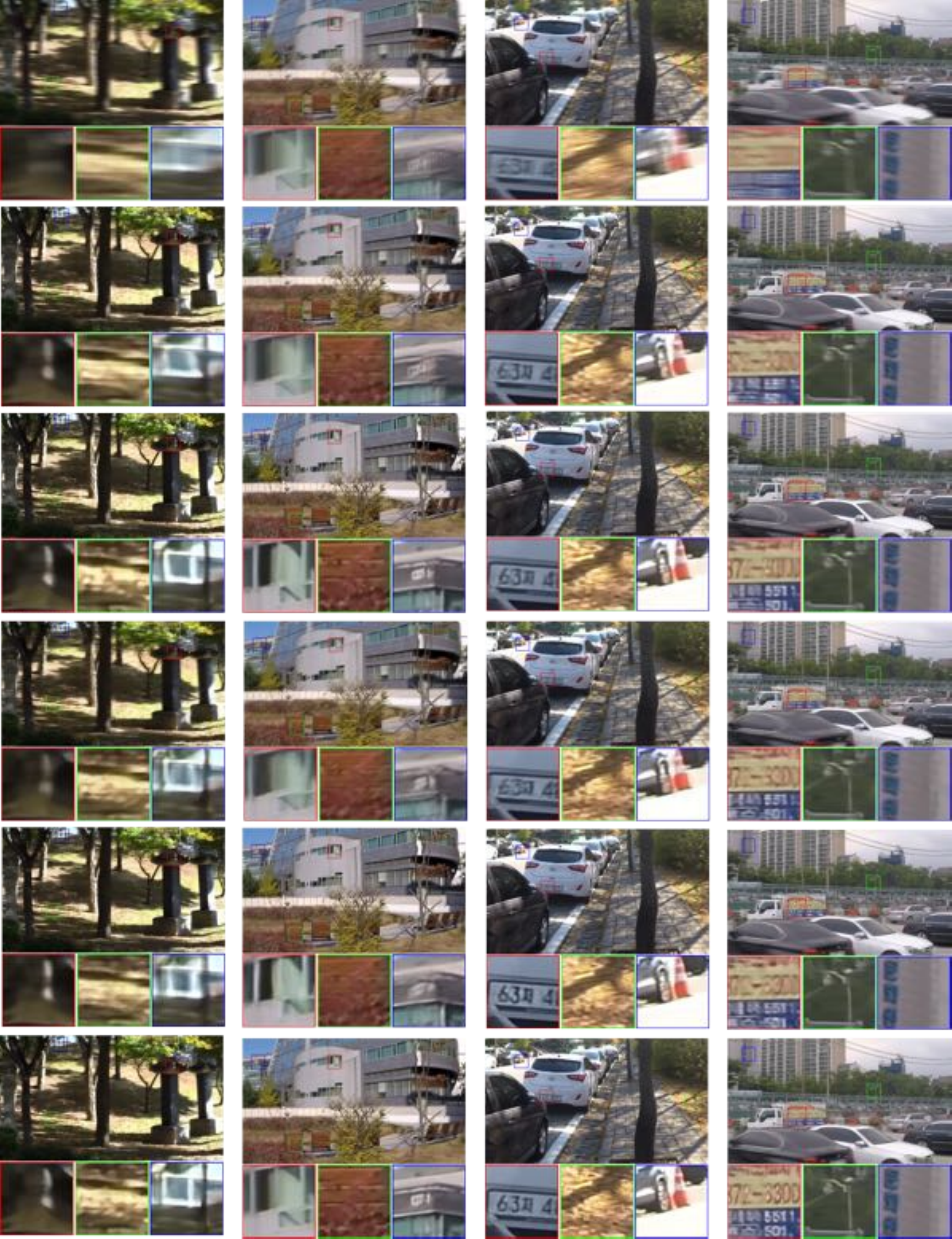}
\end{center}
\caption{Deblurring results on GoPro dataset \cite{nah2017deep} by different methods. In the top-down order, we show inputs, results of Nah \etal \cite{nah2017deep}, Tao \etal \cite{tao2018scale}, Kupyn \etal \cite{kupyn2017deblurgan}, Zhang \etal \cite{zhang2018dynamic}, and our results.}
\label{fig:results1}
\end{figure*}

%
\begin{figure*}
\footnotesize
\centering
\subfigure{
\begin{minipage}{0.333\textwidth}
\centering
\includegraphics[width=1\textwidth]{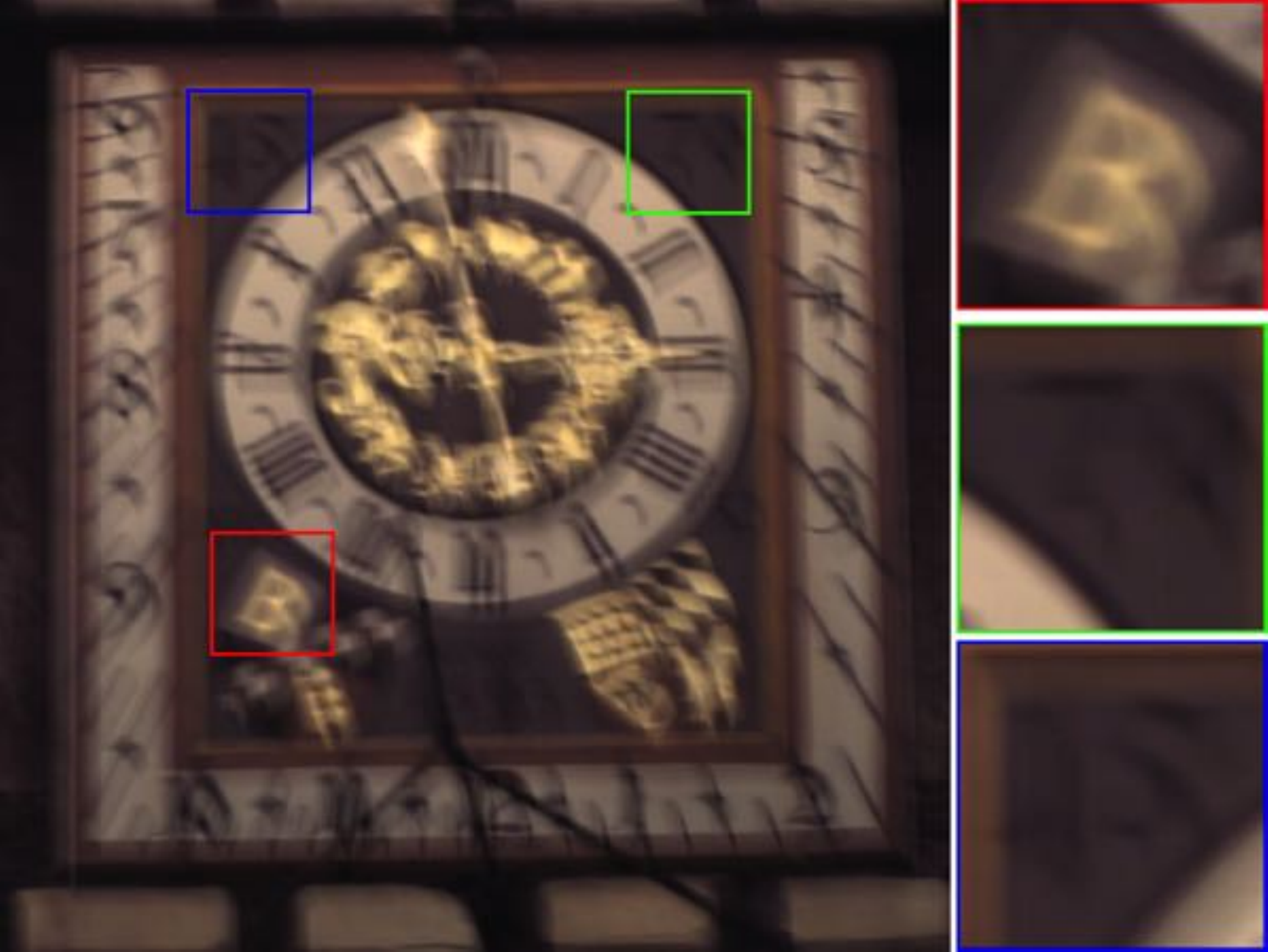}
{(a) Blurry image}
\end{minipage}
\begin{minipage}{0.333\textwidth}
\centering
\includegraphics[width=1\textwidth]{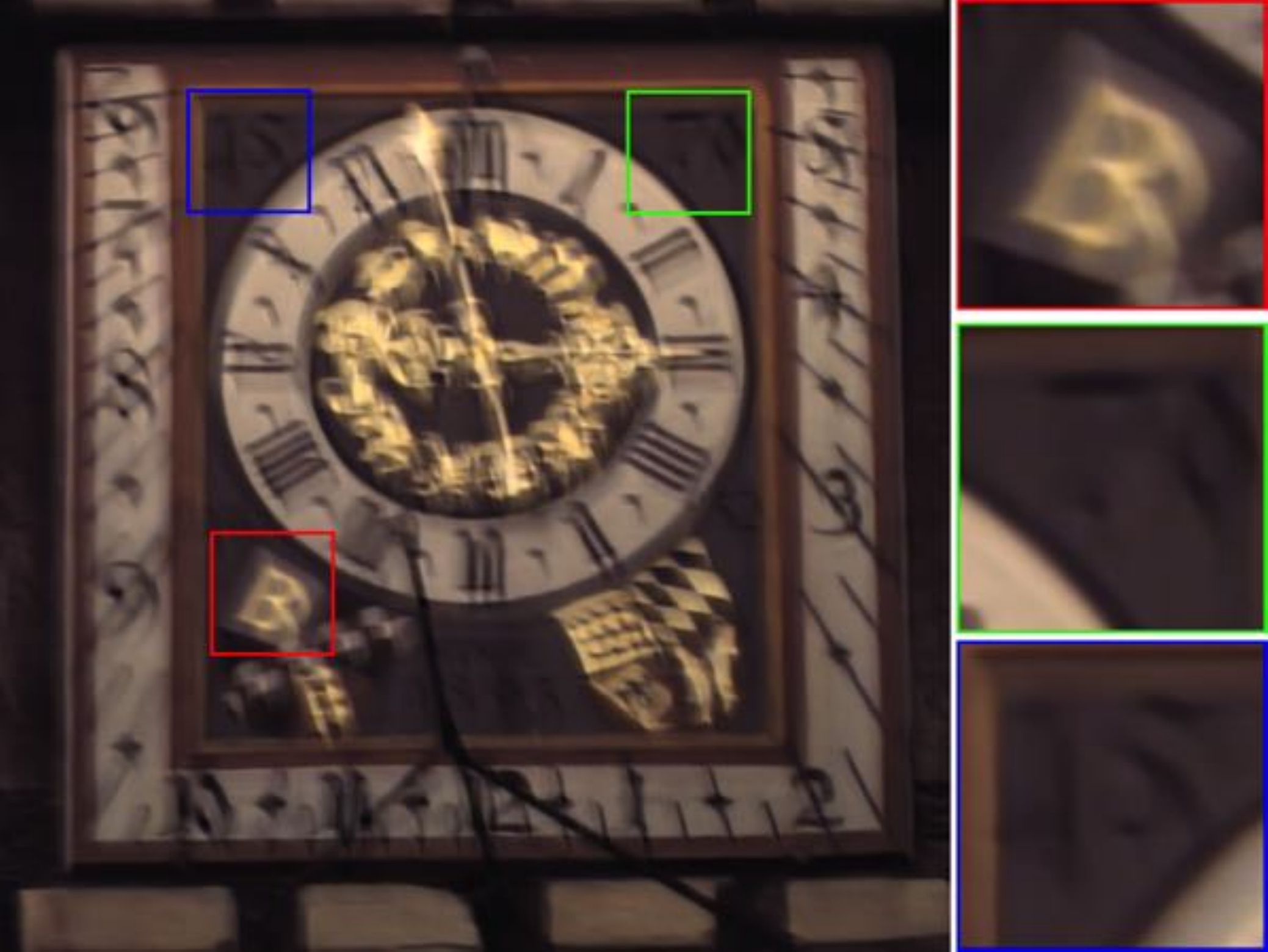}
{(b) Nah \etal \cite{nah2017deep}}
\end{minipage}
\begin{minipage}{0.333\textwidth}
\centering
\includegraphics[width=1\textwidth]{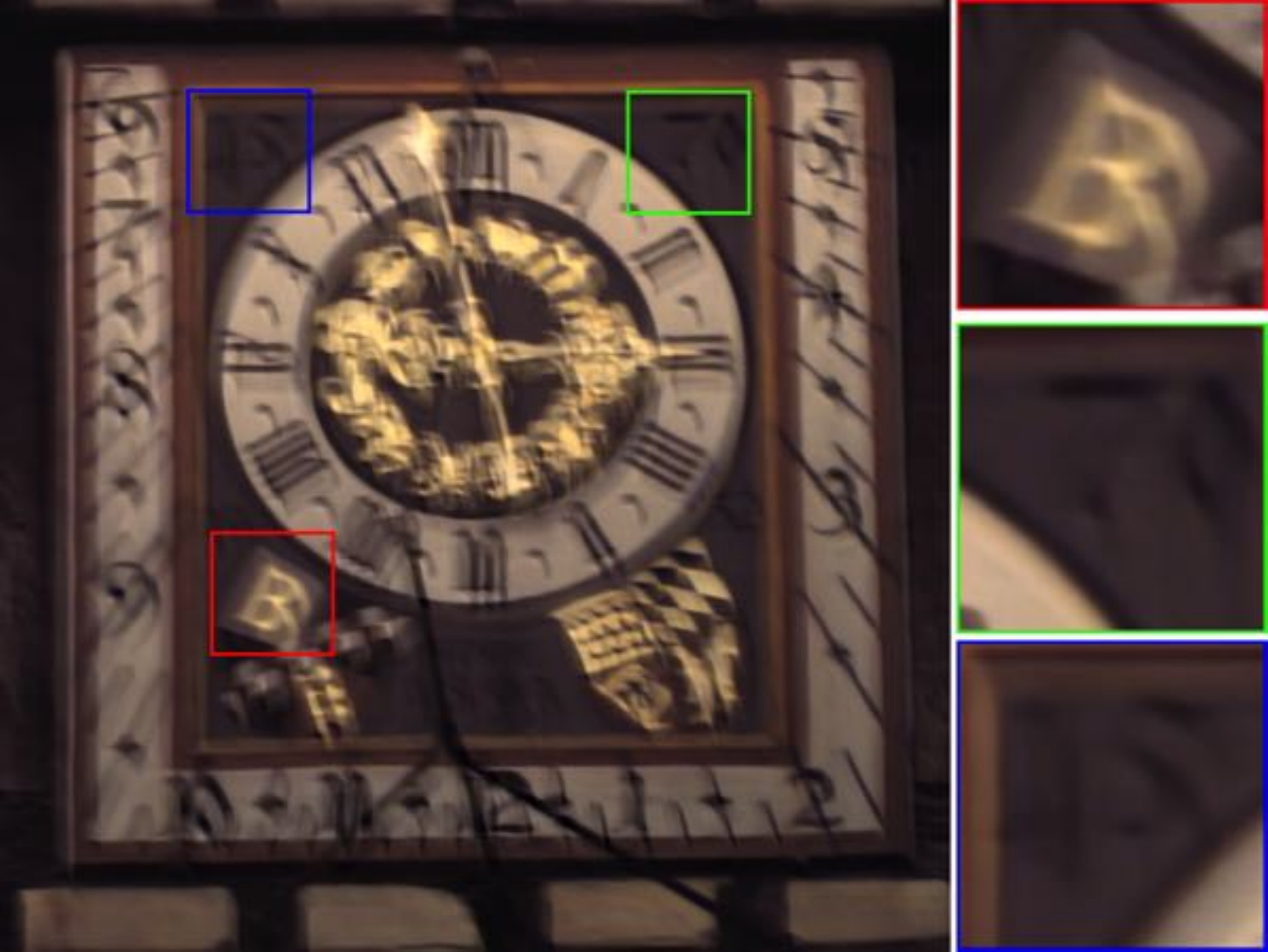}
{(c) Tao \etal \cite{tao2018scale}}
\end{minipage}
}
\subfigure{
\begin{minipage}{0.333\textwidth}
\centering
\includegraphics[width=1\textwidth]{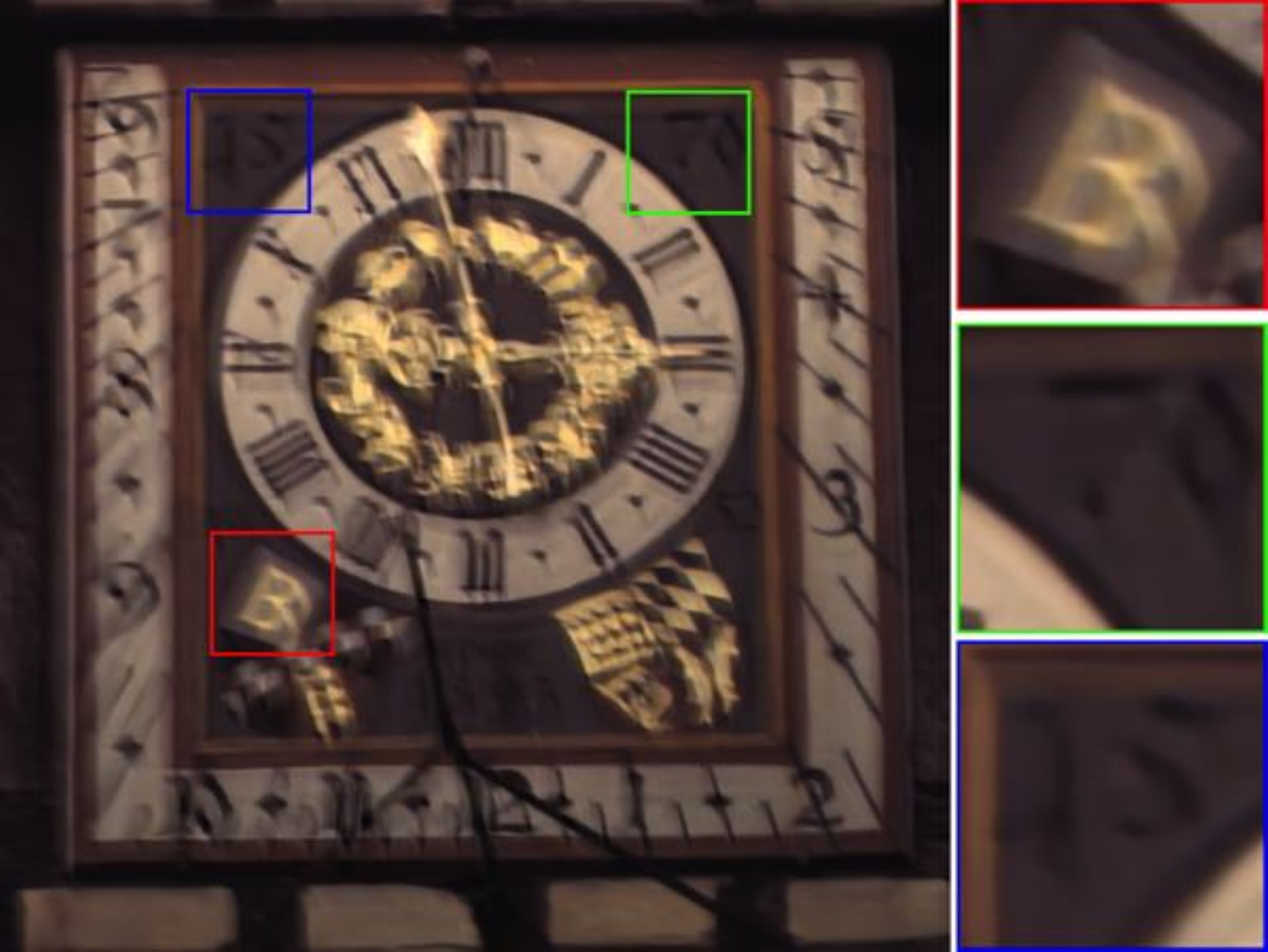}
{(d) Kupyn \etal \cite{kupyn2017deblurgan}}
\end{minipage}
\begin{minipage}{0.333\textwidth}
\centering
\includegraphics[width=1\textwidth]{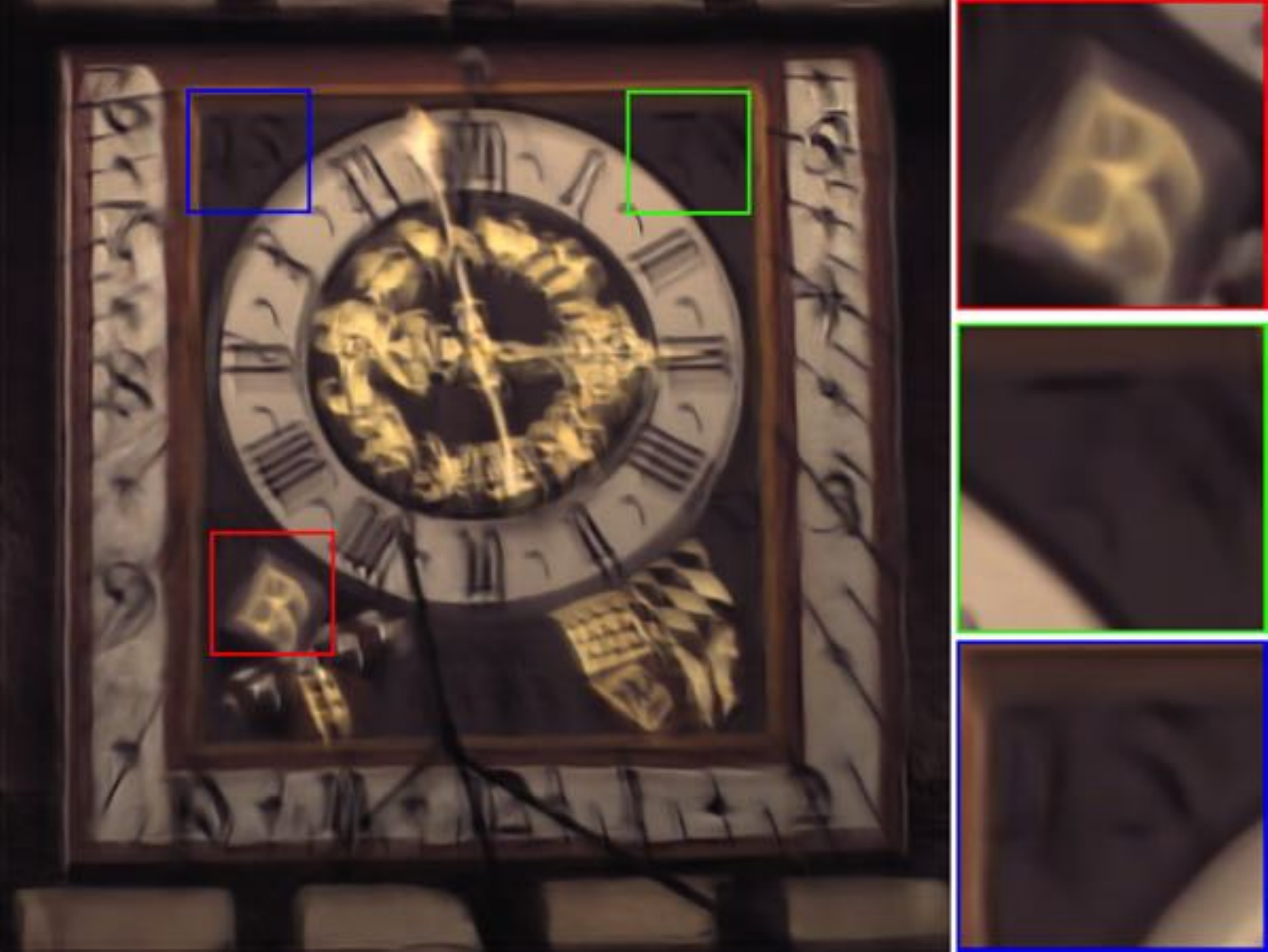}
{(e) Zhang \etal \cite{zhang2018dynamic}}
\end{minipage}
\begin{minipage}{0.333\textwidth}
\centering
\includegraphics[width=1\textwidth]{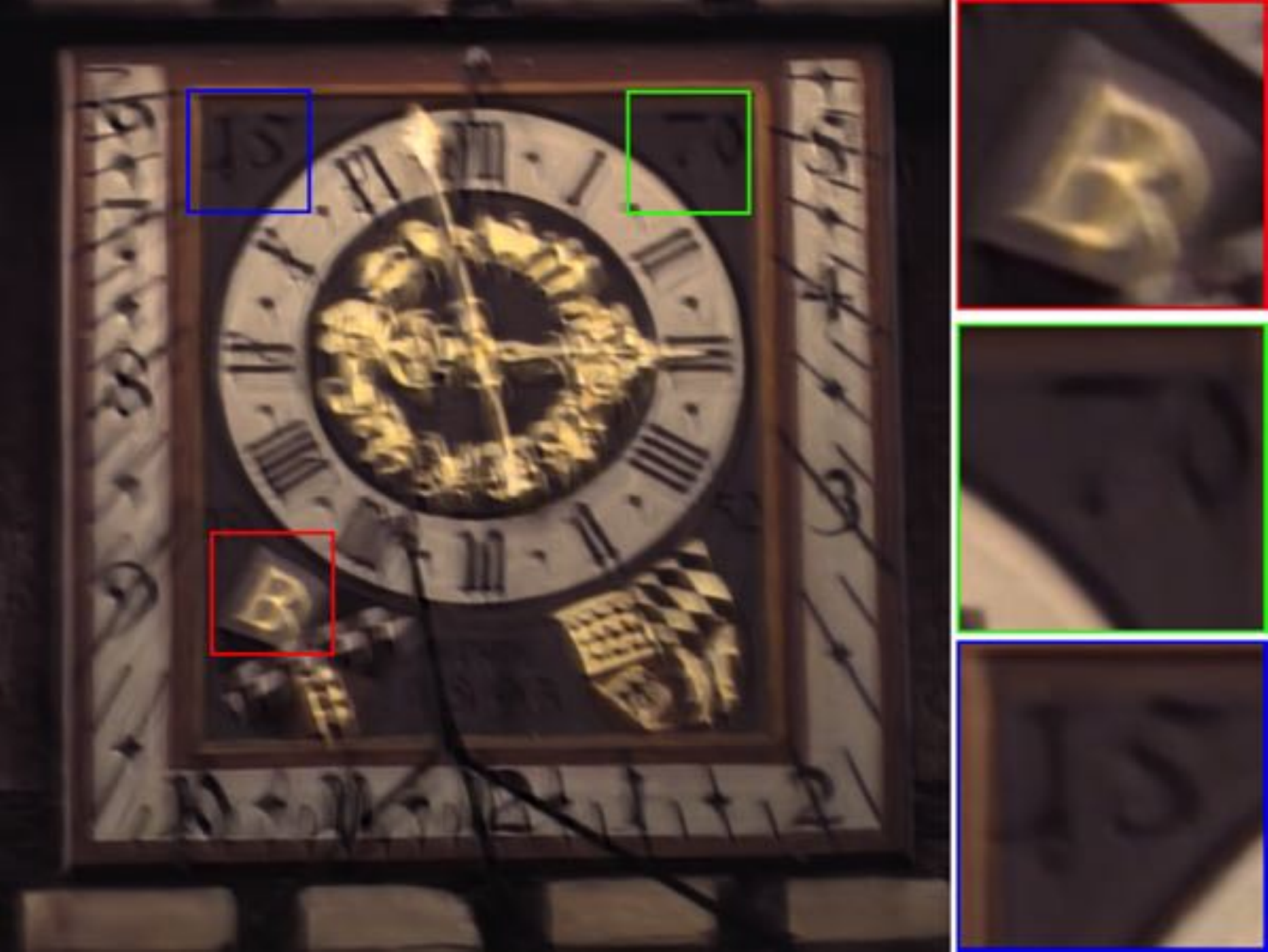}
{(f) Proposed}
\end{minipage}
}
\caption{Deblurring results on K{\"o}hler dataset \cite{kohler2012recording} by different methods.}
\label{fig:results2}
\end{figure*}
\subsection{Comparisons with State-of-the-art Methods}
In this subsection, both quantitative and qualitative evaluations are conducted to verify the proposed ECPeNet on the benchmark datasets.
\vspace{1mm}
\noindent \textbf{Quantitative Evaluations.}
We first compare the proposed ECPeNet with previous state-of-the-art deblurring methods \cite{hyun2014segmentation, sun2015learning, nah2017deep, kupyn2017deblurgan, zhang2018dynamic, tao2018scale} in a quantitative way.
The source codes and trained models of the aforementioned methods are publicly available on the authors' websites, except for \cite{hyun2014segmentation} and \cite{zhang2018dynamic} whose results have been reported in previous works \cite{nah2017deep} and \cite{zhang2018dynamic}, respectively.
Additionally, we utilize the same training dataset to retrain the network provided by \cite{zhang2018dynamic} for its evaluation on K{\"o}hler dataset.
The average PSNR, SSIM, and MSSIM indices for different deblurring methods on GoPro testing and K{\"o}hler datasets are shown in Table \ref{tab:2}.
One can see that on the GoPro testing dataset, the proposed ECPeNet significantly outperforms both the conventional non-uniform deblurring method \cite{hyun2014segmentation} and these recent developed CNN based methods \cite{sun2015learning, nah2017deep, kupyn2017deblurgan, zhang2018dynamic, tao2018scale}.
Even compared to the previous state-of-the-art method \cite{tao2018scale}, the proposed ECPeNet still has a $0.84$ dB lead.
While on the K{\"o}hler dataset, although the performance of these dynamic scene deblurring networks is comparable, it can be found that our method still has a slight advantage. 
Meanwhile, the running time by different methods for processing an image of resolution $1280\times720\times3$ is also listed in Table \ref{tab:2}.
One can notice that it takes plenty of time for a conventional method to restore an image because of the 
time-consuming iterative inference and the CPU implementation.
While for these end-to-end training networks, they can achieve much faster speed to process an image on GPU.
Considering that these dynamic scene deblurring networks are implemented by different deep learning platforms, the minor difference between them can be neglected within the margin of error.
\vspace{1mm}
\noindent \textbf{Qualitative Evaluations.}
We further compare the visual quality of restored images by our proposed ECPeNet and these recent developed CNN based dynamic scene deblurring networks, including Nah \cite{nah2017deep}, Tao \cite{tao2018scale}, Kupyn \cite{kupyn2017deblurgan}, and Zhang \cite{zhang2018dynamic}.
Figure \ref{fig:results1} shows several blurred images from the GoPro \cite{nah2017deep} testing dataset and their corresponding deblurring results produced by the above methods.
One can see that although these recent developed CNNs could remove the overall motion blur artifacts, the results restored by them are not pleasant enough because of the blurred edges and noticeable artifacts.
For example, in the fourth column, all these previous CNN based deblurring networks could not recover the text information (see the red box zoom-in region).
While in the second column, the noticeable artifacts exist around the license plate number.
By contrast, benefiting from the extreme channel prior constraint, our method can deliver a more visual pleasing result with much fewer artifacts and sharper edges.
To further demonstrate the robustness of our method, the visual comparison results on images from the K{\"o}hler \cite{kohler2012recording} dataset are also provided in Figure \ref{fig:results2}.
Again, it can be seen that artifacts and blurred edges in the zoom-in areas (see characters `$B$', `$70$', and `$15$') are noticeable for these previous CNN based methods.  
Although results recovered by Kupyn \cite{kupyn2017deblurgan} and Tao \cite{zhang2018dynamic} are sharper than other methods, distortion still exists.
Compared with these methods, the ECPeNet can restore image sharpness and naturalness.
\section{Conclusion}
In this work, we presented a simple yet effective Extreme Channel Prior embedded Network (ECPeNet) with a novel trainable extreme channel prior embedded layer (ECPeL), which aims to integrate extreme (\ie, dark and bright) channel priors into a deep CNN for dynamic scene deblurring. 
By extracting the extreme channels of shallow features and enforcing sparsity on them, ECPeNet can regularize the solution space of the network. 
Additionally, ECPeNet works in both coarse-to-fine and fine-to-coarse manners to exploit information of blurred images at different resolutions to maximize information flow across scales.
Benefiting from the extreme channel prior constraint and effective multi-scale network architecture, the developed ECPeNet outperforms previous dynamic scene deblurring networks by a large margin.
Quantitative evaluations on the challenging GoPro dataset showed that the proposed ECPeNet had at least $0.84$ dB PSNR gains over the existing state-of-the-arts.
%

%
%

{\small
\bibliographystyle{ieee}
\bibliography{egbib}
}

\end{document}